\documentclass[conference]{IEEEtran}
\IEEEoverridecommandlockouts
\usepackage{cite}
\usepackage{amsmath,amssymb,amsfonts}
\usepackage{algorithmic}
\usepackage{graphicx}
\usepackage{textcomp}
\usepackage{xcolor}
\def\BibTeX{{\rm B\kern-.05em{\sc i\kern-.025em b}\kern-.08em
    T\kern-.1667em\lower.7ex\hbox{E}\kern-.125emX}}

\usepackage[caption=false]{subfig}
\usepackage{textcomp}
\usepackage{stfloats}
\usepackage{url}
\usepackage{verbatim}
\usepackage{graphicx}
\usepackage{cite}

\usepackage[english]{babel}
\usepackage{soul,color}
\soulregister\cite7
\soulregister\ref7
\soulregister\ac7
\soulregister\acp7
\soulregister\underline7

\usepackage{threeparttable}
\usepackage{url}
\usepackage{cite}
\DeclareMathAlphabet{\pazocal}{OMS}{zplm}{m}{n}
\usepackage{nicefrac}
\usepackage{optidef}
\usepackage{gensymb}
\usepackage{xurl}
\usepackage{booktabs}
\usepackage{xcolor}

\begin{document}

\title{Learning to Focus: CSI-Free Hierarchical MARL for Reconfigurable Reflectors\\
% {\footnotesize \textsuperscript{*}Note: Sub-titles are not captured for https://ieeexplore.ieee.org  and
% should not be used}
\thanks{This material is based upon work supported in part by the U.S. Department of Energy, Office of Science, Office of Advanced Scientific Computing Research, Early Career Research Program under Award Number DE-SC0023957, and in part by the National Science Foundation under Grant No. 2549088.}
}

% \author{\IEEEauthorblockN{1\textsuperscript{st} Given Name Surname}
% \IEEEauthorblockA{\textit{dept. name of organization (of Aff.)} \\
% \textit{name of organization (of Aff.)}\\
% City, Country \\
% email address or ORCID}
% \and
% \IEEEauthorblockN{2\textsuperscript{nd} Given Name Surname}
% \IEEEauthorblockA{\textit{dept. name of organization (of Aff.)} \\
% \textit{name of organization (of Aff.)}\\
% City, Country \\
% email address or ORCID}
% \and
% \IEEEauthorblockN{3\textsuperscript{rd} Given Name Surname}
% \IEEEauthorblockA{\textit{dept. name of organization (of Aff.)} \\
% \textit{name of organization (of Aff.)}\\
% City, Country \\
% email address or ORCID}
% \and
% \IEEEauthorblockN{4\textsuperscript{th} Given Name Surname}
% \IEEEauthorblockA{\textit{dept. name of organization (of Aff.)} \\
% \textit{name of organization (of Aff.)}\\
% City, Country \\
% email address or ORCID}
% \and
% \IEEEauthorblockN{5\textsuperscript{th} Given Name Surname}
% \IEEEauthorblockA{\textit{dept. name of organization (of Aff.)} \\
% \textit{name of organization (of Aff.)}\\
% City, Country \\
% email address or ORCID}
% \and
% \IEEEauthorblockN{6\textsuperscript{th} Given Name Surname}
% \IEEEauthorblockA{\textit{dept. name of organization (of Aff.)} \\
% \textit{name of organization (of Aff.)}\\
% City, Country \\
% email address or ORCID}
% }

\author{\IEEEauthorblockN{Hieu Le}
\IEEEauthorblockA{\textit{Electrical and Computer Engineering} \\
\textit{Texas A\&M University}\\
College Station, Texas, USA \\
hieult@tamu.edu} \\

\\
\IEEEauthorblockN{Jian Tao}
\IEEEauthorblockA{\textit{School of Performance, } \\
\textit{Visualization, and Fine Arts} \\
\textit{Texas A\&M University}\\
College Station, Texas, USA \\
jtao@tamu.edu }
\and
\IEEEauthorblockN{Mostafa Ibrahim}
\IEEEauthorblockA{\textit{Engineering Technology } \\
\textit{and Industrial Distribution} \\
\textit{Texas A\&M University}\\
College Station, Texas, USA \\
mostafa.ibrahim@tamu.edu }
\\
\IEEEauthorblockN{Sabit Ekin}
\IEEEauthorblockA{\textit{Engineering Technology} \\
\textit{and Industrial Distribution} \\
\textit{Texas A\&M University}\\
College Station, Texas, USA \\
sabitekin@tamu.edu }
\and
\IEEEauthorblockN{Oguz Bedir}
\IEEEauthorblockA{\textit{Electrical and Computer Engineering} \\
\textit{Texas A\&M University}\\
College Station, Texas, USA \\
oguzbedir@tamu.edu }
}

\maketitle

\begin{abstract}

Reconfigurable Intelligent Surfaces (RIS) have the potential to engineer smart radio environments for next-generation millimeter-wave (mmWave) networks. However, the prohibitive computational overhead of Channel State Information (CSI) estimation and the dimensionality explosion inherent in centralized optimization severely hinder practical large-scale deployments. To overcome these bottlenecks, we introduce a per-element CSI-free paradigm powered by a Hierarchical Multi-Agent Reinforcement Learning (HMARL) architecture to control mechanically reconfigurable reflective surfaces. By substituting pilot-based channel estimation for each element of the device with accessible user localization data, our framework leverages spatial intelligence for macro-scale wave propagation management. The control problem is decomposed into a two-tier neural architecture: a high-level controller executes temporally extended, discrete user-to-reflector allocations, while low-level controllers autonomously optimize continuous focal points using Multi-Agent Proximal Policy Optimization (MAPPO) under a Centralized Training with Decentralized Execution (CTDE) scheme. Comprehensive deterministic ray-tracing evaluations in an indoor mmWave scenario demonstrate that this hierarchical framework achieves received signal strength indicator (RSSI) improvements of up to 7.79 dB over centralized Proximal Policy Optimization (PPO) baselines. Furthermore, the system maintains resilient beam-focusing performance under practical sub-meter localization tracking errors for up to four users and two reflector arrays. By eliminating execution-time CSI overhead while preserving high-fidelity signal redirection, this work provides a scalable and cost-effective step toward intelligent indoor wireless environments.
\end{abstract}


\section{Introduction}
\label{introduction}

The unprecedented surge in wireless traffic demand has driven conventional communication architectures to their theoretical limits \cite{direnzo:2020}. 
In response, reconfigurable intelligent surfaces (RIS) have emerged as a transformative technology, turning the previously passive radio propagation medium into a dynamic, controllable environment. 
Conventional RIS architectures leverage electronically controlled phase shifters to induce constructive interference at receivers \cite{direnzo:2020, zahra:2021}.
However, these systems depend critically on accurate channel state information (CSI) for each reflecting unit \cite{basharat:2022}. 
As deployments scale to hundreds of elements, the pilot overhead required for cascaded channel estimation becomes a prohibitive computational bottleneck, causing high spectral efficiency loss \cite{a9400843}.

While deep reinforcement learning (DRL) and multi-agent reinforcement learning (MARL) have been increasingly adopted to address the optimization complexity of RIS coordination, most existing frameworks still mandate explicit channel estimation \cite{huang2020reconfigurable, taha2020deep, taha2021enabling}. 
Efforts to relax this CSI dependence often require massive offline training datasets or the integration of dedicated sensing hardware into the RIS, which substantially increases system cost, power consumption, and hardware complexity \cite{choi2024deep, sheen2021deep}.

To circumvent these fundamental limitations, we shift focus from electronic metasurfaces to mechanically reconfigurable metallic reflectors. 
Unlike electronic architectures that require complex RF circuitry, metallic reflectors offer inherent wideband operation and simplified mechanical actuation \cite{le2024guiding, a8972365, aa11322690}. 

In our prior work, we established the physical foundation and control viability of this mechanical approach. We demonstrated that arrays of metallic flat reflectors, acting as linear Fresnel reflectors, can provide substantial coverage and gain enhancements in non-line-of-sight (NLOS) environments, offering a low-cost, frequency-versatile alternative to electronic phase-shifters \cite{le2024guiding}. Building upon this hardware design, we introduced a MARL framework to guide these reflector arrays, demonstrating that distributed control outperforms centralized DRL baselines in multi-user scenarios \cite{aa11322690}. However, while standard MARL mitigates the dimensionality explosion of centralized approaches, coordinating simultaneous discrete user allocation and continuous beam-focusing across massive multi-reflector environments remains a formidable computational challenge.

To overcome this remaining complexity bottleneck while leveraging the mechanical hardware, we propose a fundamentally different methodology that eliminates per-element pilot-based electromagnetic channel estimation. Instead of coordinating fine-grained electromagnetic interference, our framework exploits spatial awareness and readily available user localization data to manage macro-scale signal propagation in these NLOS environments.

While replacing per-element CSI with localization information introduces a dependency on 
continuous spatial tracking, acquiring high-fidelity, decimeter-level user 
context is highly feasible and cost-effective in modern wireless networks. 
Recent advancements in joint communication and sensing systems have 
demonstrated that commercial off-the-shelf infrastructure can provide 
continuous, high-update-rate localization without necessitating expensive, 
dedicated tracking hardware. For instance, integrated millimeter-wave radar 
systems have been shown to achieve robust sub-decimeter localization accuracy 
in dense indoor environments \cite{zhang2023push}. Furthermore, ubiquitous 
Wi-Fi infrastructure can now be leveraged to provide highly accurate spatial 
tracking and orientation data \cite{chi2022widrone}. By relying on these 
increasingly standard sensing modalities, our framework shifts 
the system overhead from per-element electromagnetic channel 
estimation to spatial localization, which scales far more efficiently 
in multi-user deployments.

To manage the massive combinatorial complexity of joint user assignment and continuous control, we formulate the problem as a hierarchical multi-agent reinforcement learning (HMARL) framework \cite{makar2001hierarchical}. The architecture decomposes the task into two temporal abstraction levels: a centralized high-level controller that performs discrete user-to-reflector allocations, and low-level controllers that autonomously optimize continuous focal points for their assigned users. Trained using multi-agent proximal policy optimization (MAPPO) \cite{yu2022surprising} under a centralized training with decentralized execution (CTDE) paradigm \cite{kraemer2016multi}, this decomposition ensures rapid learning and practical deployment scalability. Furthermore, the practical deployment viability is highly supported by the widespread availability of specialized deep learning accelerators in modern communication systems \cite{nasari2022benchmarking, le2024insight}. Building on our prior mechanically reconfigurable reflector studies, which focused on single-level DRL control and coverage enhancement \cite{le2024guiding, aa11322690}, this work introduces an explicit hierarchical decomposition and a policy that rely on user localization and geometric priors rather than per-element CSI.

In this paper, we demonstrate the efficacy of this per-element CSI-free paradigm.
\begin{itemize}
\item \textbf{Per-element CSI-free optimization:} A fully functional HMARL framework that eliminates execution-time CSI estimation overhead, utilizing spatial localization and a pre-calibrated ray-tracing model to achieve significant received signal strength indicator (RSSI) improvements over centralized PPO baselines.
\item \textbf{Hardware and algorithmic analysis:} Comprehensive ray-tracing-based validation demonstrating resilient beam focusing under practical sub-meter indoor localization errors and mechanically feasible focal-point adjustments.
\end{itemize}

\section{System Model and Problem Formulation}
\label{sec:system_model}

\subsection{Mechanically Reconfigurable Reflector \& Hierarchical Coordination}

\begin{figure}[!t]
    \centering
    \includegraphics[width=0.9\linewidth]{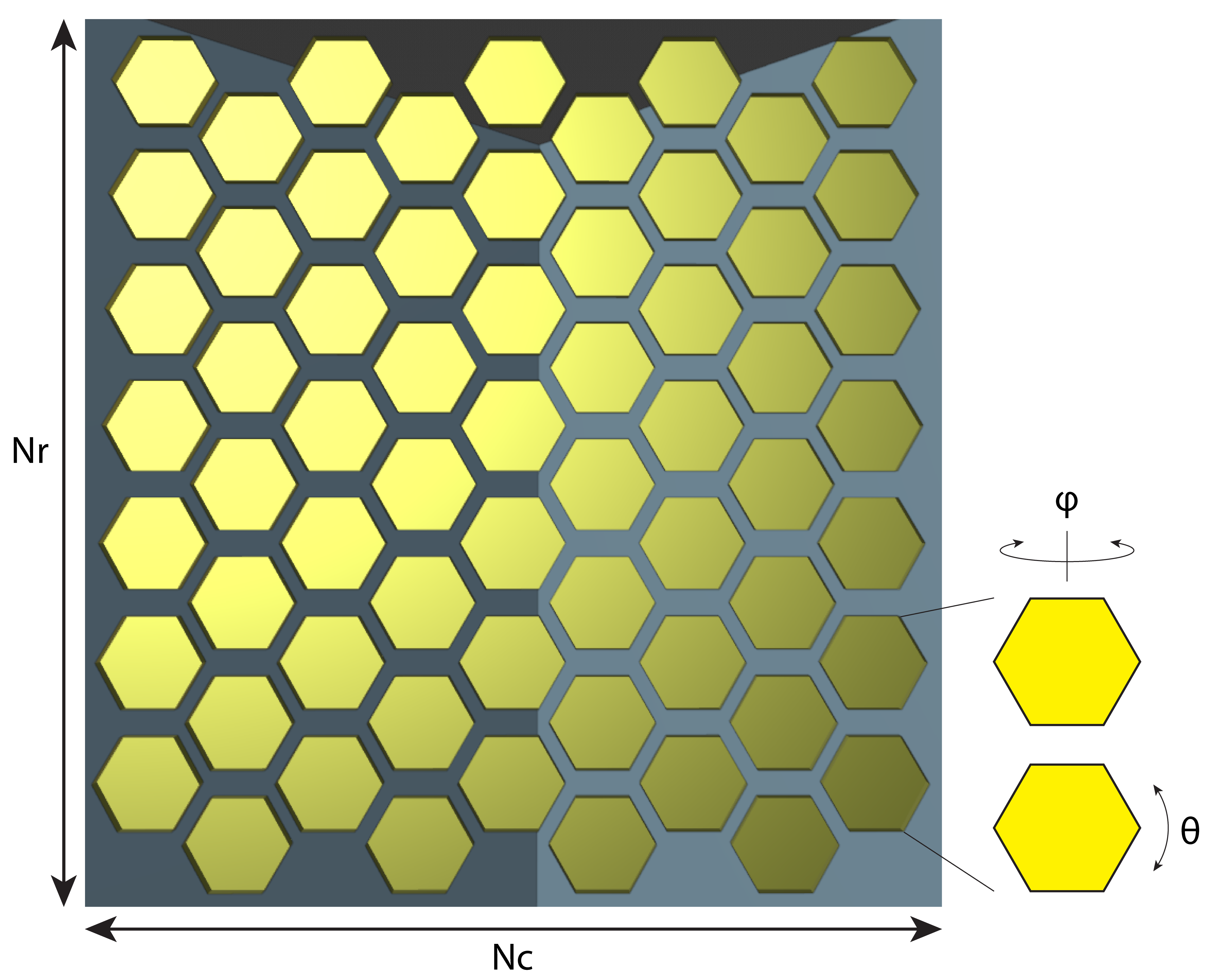}
    \caption{Reflector Design.}
    \label{Figure:reflector}
\end{figure}

The system comprises an access point (AP) located at $s \in \mathbb{R}^3$, $K$ user equipment (UE) devices, and $L$ independent reflector segments operating in a NLOS millimeter-wave (mmWave) environment. 
Unlike standard electronic phase-shifters, the reflector is a mechanical device consisting of many small metallic tiles arranged in an $N_r \times N_c$ grid (Fig.~\ref{Figure:reflector}).

To bypass the severe computational bottleneck of per-tile optimization, we optimize for beam-focusing with a focal point so that the reflector's tiles focus energy toward that point. 
For a focal point $f_l(t) \in \mathbb{R}^3$ associated with reflector segment $l$, the mechanical orientation of tile $(i, j)$ at position $r_{i,j}$ is deterministically governed by its normal vector:
\begin{equation}
\vec{n}_{i,j}(f_l) = \frac{1}{2} \left( 
\frac{f_l - r_{i,j}}{\|f_l - r_{i,j}\|_2} + 
\frac{s - r_{i,j}}{\|s - r_{i,j}\|_2} 
\right).
\label{eq:normal_vector}
\end{equation}

This geometric formulation allows us to derive the necessary elevation $\theta_{i,j}$ and azimuth $\phi_{i,j}$ angles without requiring instantaneous electromagnetic CSI.

To manage the massive combinatorial complexity of a multi-user, multi-reflector environment, we decompose the control problem into a two-tier HMARL architecture. 
There are two levels: a high-level user–reflector assignment controller and low-level reflector focal point control agents. 
Operating at an extended timescale $T$, the high-level controller observes the global spatial state to determine the discrete user-to-reflector allocation $b = \{b_1, \dots, b_L\}$. 
Given this assignment, the decentralized low-level controllers autonomously execute continuous focal-point displacements
\begin{equation}
a_{l,t} = [\Delta f_{l,x}, \Delta f_{l,y}, \Delta f_{l,z}]^T
\end{equation}
at every environmental timestep to maximize the signal strength for their assigned users.

\subsection{Signal Propagation \& Optimization Objective}

In the considered NLOS mmWave environment, the direct path between the AP and the users is assumed to be obstructed. Consequently, the controllable RSSI at user $k$ relies entirely on the reflected paths facilitated by its assigned reflector segment. It is noted that we use RSSI as a proxy for link quality, defined here as the received power at each user computed from deterministic ray-tracing, expressed in dBm. For a user $k$ assigned to segment $l$ under the high-level allocation $b$, the received power is formulated as:
\begin{equation}
P_{r,k}(u_k, f) = 
P_t \sum_{(i,j) \in \mathcal{S}_{b_l}} 
\left| 
h_{r,k}^{(i,j)}(u_k, f) + h_{other,k}(u_k)
\right|^2,
\label{eq:received_power}
\end{equation}
where $P_t$ is the transmit power, $h_{r,k}^{(i,j)}(u_k, f)$ represents the reflected channel coefficient from tile $(i, j)$ as a function of the user, $u_k$, and focal point, $f$, locations, and $h_{other,k}(u_k)$ accounts for other environmental propagation paths. 
Crucially, these coefficients are derived from deterministic ray tracing of a fixed propagation environment based on user and focal-point localization.

The system objective is to maximize the aggregate received power across all $K$ users. 
We define the system-wide performance metric at time step $t$ as a state-dependent reward function:
\begin{equation}
R_{sys}(s(t), b(t)) = 
\sum_{k=1}^{K} P_{r,k}(u_k(t), f(t)).
\label{eq:system_reward}
\end{equation}
This formulation establishes a differentiable performance metric that directly couples the discrete high-level allocation decisions $b(t)$ with the continuous low-level focal point configurations $f_l(t)$, thus enabling coordinated hierarchical optimization.

By elevating the optimization space from individual tile orientations to segment-level focal points, the control parameter dimension is reduced from
\begin{equation}
\mathcal{D}_{\text{tile}} = K^L + 2 N_r N_c
\end{equation}
to a highly compact representation of
\begin{equation}
\mathcal{D}_{\text{focal}} = K^L + 3L.
\end{equation}
For dense indoor deployments where the hardware complexity term $2N_r N_c$ typically dominates the segment count term $3L$, this dimensionality reduction yields the fundamental computational feasibility required for fast MARL convergence.

\section{Hierarchical MARL Framework}

\begin{figure}[!tbp]
    \centering
    \includegraphics[width=\columnwidth]{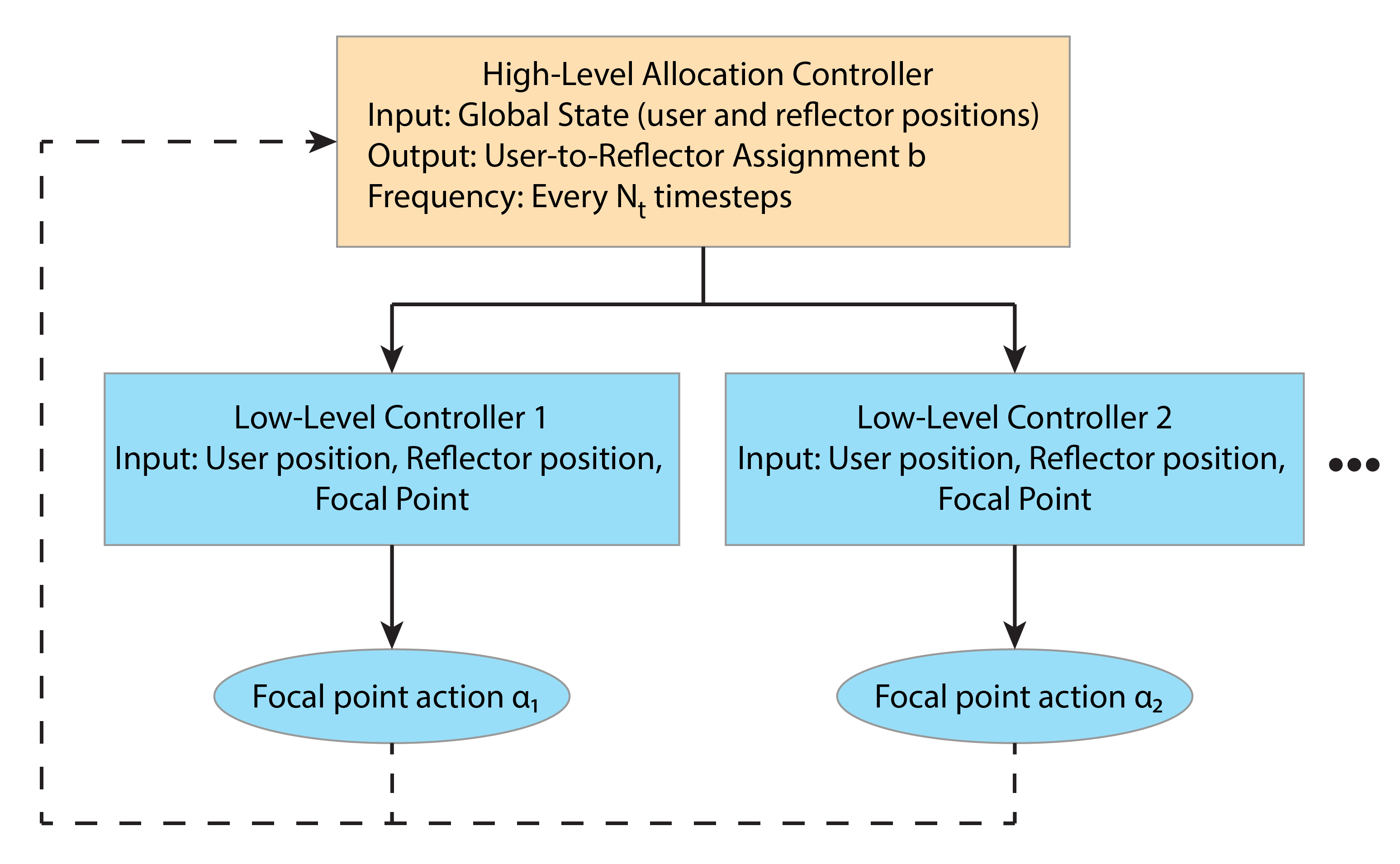} 
    \caption{Hierarchical Multi-Agent Reinforcement Learning Architecture. The high-level controller evaluates the global system state to determine discrete user-to-reflector allocations every $T$ timesteps. Concurrently, low-level agents utilize masked local observations to continuously optimize focal points using PPO under a CTDE scheme.}
    \label{fig:workflow_architecture}
\end{figure}

To solve the joint optimization of user assignment and focal point placement, we formulate the problem as a Hierarchical Multi-Agent Markov Decision Process (HMA-MDP). 
For the HMARL, there are two distinct control levels: a high-level user–reflector assignment controller and decentralized low-level reflector focal point controllers (Fig.~\ref{fig:workflow_architecture}.)

\subsection{Hierarchical Coordination Architecture}

\subsubsection{High-Level Allocation Controller}
Operating as a centralized decision-maker, the high-level controller observes the global system state
\begin{equation}
s_H(t) = \{u_1(t), \dots, u_K(t), r_1, \dots, r_L, f_1(t), \dots, f_L(t)\},
\end{equation}
to systematically evaluate user-to-reflector assignments. 
The controller selects a discrete combinatorial action $b(t) \in \mathcal{B}$ with cardinality $|\mathcal{B}| = K^L$. 
To ensure learning stability, this level operates with temporal abstraction, the controller only updates its allocation every $T$ environmental timesteps. 
This extended timescale provides a stable optimization horizon, allowing the low-level controllers sufficient time to adapt their focal points before reassignment occurs, thereby preventing destructive policy oscillations.

\subsubsection{Low-Level Focal Point Agents}
Given an allocation $b$ from the controller, each reflector segment $l$ acts as an independent agent. 
To ensure multi-agent scalability and decouple the learning process, strict observation masking is applied.
With $\pi(l)$ which allocates reflector segment to each user, 
each agent $l$ executes its policy based exclusively on a localized observation.
\begin{equation}
o_{L,l}(t) = \{u_{\pi(l)}(t), r_l, f_l(t)\},
\end{equation}
which contains only the position of its assigned user, its own reflector position, and its current focal point. 
The agent continuously outputs displacement actions $a_{l,t} \in \mathbb{R}^3$ at every timestep, bounded by a maximum mechanical actuation limit $\delta_{\max}$. 
This localized execution reduces each agent's observation space dimensionality from $\mathbb{R}^{3K + 6L}$ to a reduced dimensional $\mathbb{R}^9$.

\subsubsection{MAPPO with CTDE}
We optimize the policies using MAPPO governed by the CTDE paradigm. 
During centralized training, a global critic network evaluates the complete system state $s_{\text{H}}$ to compute accurate advantage estimates, effectively resolving the non-stationarity inherent in concurrent multi-agent updates. 
During deployment, the global critic is discarded, allowing the low-level controllers to execute optimal focal point adjustments purely through decentralized local observations without any inter-agent communication overhead.

\subsection{MAPPO with CTDE and Compatibility Matrix}

Both the high-level controller and the low-level focal point controllers are trained using MAPPO. 
To ensure stable cooperative learning, we employ a CTDE scheme. 
During the centralized training phase, a global critic evaluates the joint state $s_{\text{H}, t}$ to compute the advantage function $Adv(s_{\text{H},t}, a_t^l)$, which provides accurate credit assignment and mitigates the multi-agent non-stationarity problem \cite{yu2022surprising}. 
Each agent $l$ then updates its policy network by maximizing the clipped surrogate objective:
\begin{equation}
\begin{aligned}
\mathcal{L}^{\text{CLIP}}(\psi_l) = \mathbb{E}_t \Big[ 
\min \Big( 
r_t^l(\psi_l) \, Adv,\, \\
\mathrm{clip}\big(r_t^l(\psi_l), 1-\epsilon, 1+\epsilon\big) \, Adv 
\Big) 
\Big],
\end{aligned}
\end{equation}
where $r_t^l(\psi_l)$ denotes the probability ratio of the action under the current and previous policies, $r_t^l(\psi_l) = \frac{\pi_{\psi_l}(a_t^l|o_t^l)}{\pi_{\psi_{\text{old}},l}(a_t^l|o_t^l)}$ with $\psi_l$ is the low-level controller agent $l$.

While the MAPPO setup ensures stable execution, the high-level allocation space still scales exponentially as $K^L$, creating a sparse reward landscape. 
Discovering near-optimal assignments through pure exploration is computationally impractical within standard training horizons. 
To accelerate convergence, we introduce a domain-specific compatibility matrix $C \in \mathbb{R}^{K \times L}$ that encodes prior geometric knowledge as an inductive bias.

The matrix element $C_{kl}$ quantifies the expected signal propagation favorability when user $k$ is assigned to reflector segment $l$:
\begin{equation}
C_{kl} = \exp\left(-\frac{\|u_k - r_l\|}{d_0}\right) \cdot \cos(\theta_{kl}),
\end{equation}
where $\|u_k - r_l\|$ is the Euclidean distance between the user and the reflector, $d_0$ is a normalization constant, and $\theta_{kl}$ is the AP–reflector–user reflection angle.

Rather than acting as simple reward shaping, this matrix serves as an inductive bias injected directly into the high-level allocation policy:
\begin{equation}
\pi_H(b \mid s_H;\phi) \propto 
\left( 
Q_H(s_H, b) + \alpha(t) \sum_{k=1}^K C_{k,b_l} 
\right).
\end{equation}
The coefficient $\alpha(t)$ acts as a temporal decay gate; it heavily weighs the geometric prior during the initial exploration phase and drops to zero once a predefined episode threshold is reached. 
This structured guidance allows the high-level controller to bypass many suboptimal combinatorial configurations, accelerating the early learning phases.

\section{Results and Discussion}
\label{sec:results}

\subsection{Simulation Setup and Concurrent Environments}

\begin{figure}[!tbp]
    \centering
    \includegraphics[width=\columnwidth]{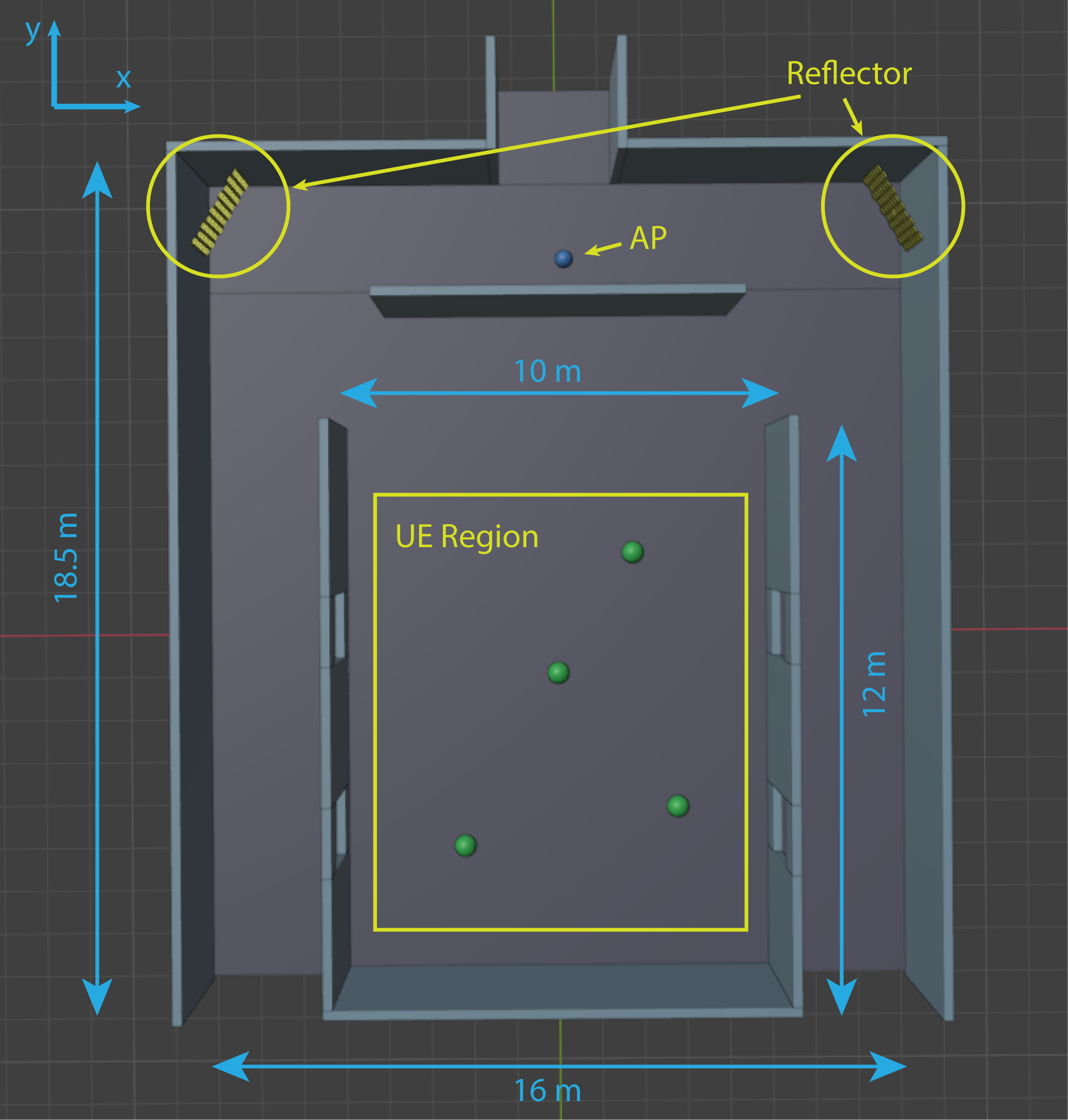}
    \caption{Experimental setup of the conference room simulation environment. The access point (AP) is positioned outside the room, serving users within the designated $10 \text{ m} \times 10 \text{ m}$ coverage region. Two mechanically reconfigurable reflectors are deployed at the corners to establish NLOS links.}
    \label{fig:simulation_setup}
\end{figure}

To empirically validate the proposed HMARL framework, we construct a high-fidelity \(60\,\text{GHz}\) indoor mmWave simulation environment. 
The experimental testbed models a conference room where an AP is positioned externally, serving $K$ users uniformly distributed within a \(10\,\text{m} \times 10\,\text{m}\) coverage area (Fig.~\ref{fig:simulation_setup}). 
Two mechanically reconfigurable metallic reflector arrays are deployed at the room's corners to establish NLOS links. 
The total transmit power is constrained to \(5\,\text{dBm}\) to represent a low-power mmWave communication system.

Electromagnetic propagation is modeled using NVIDIA Sionna's deterministic ray-tracing engine integrated with Blender. 
To ensure realistic multipath phenomena, including reflections, diffractions, and scattering, structural materials are strictly parameterized according to ITU-R P.2040-1 standards. 
This includes concrete walls (relative permittivity, \(\epsilon_r = 5.31\) and conductivity, \(\sigma = 0.0326\,\text{S/m}\)) and marble floors (\(\epsilon_r = 7.0\), \(\sigma = 0.01\,\text{S/m}\)), while the reflector tiles are modeled as highly conductive metallic panels (\(\epsilon_r = 1\)).

Given the substantial computational overhead of generating ray-tracing data for millions of continuous HMARL training steps, the simulation architecture is custom-built to leverage highly parallelized concurrent environments. 
We utilize multithreading to instantiate multiple simulation replicas simultaneously. 
Because NVIDIA Sionna is optimized for hardware acceleration, the computationally intensive ray-tracing operations are entirely offloaded to the GPU, while the CPU manages the environment logic, result gathering, and trajectory storage. 
By running different environment configurations in parallel and synchronously gathering the batch results at the end of each step, the framework achieves the massive sample throughput required for stable MAPPO convergence without bottlenecking the training pipeline.

\begin{table}[htbp]
\caption{Simulation Environment and HMARL Hyperparameters}
\label{tab:system_params}
\centering
\begin{tabular}{p{4.2cm} l}
\hline
\textbf{Parameter Description} & \textbf{Assigned Value} \\
\hline
\multicolumn{2}{c}{\textit{Environment \& Hardware Setup}} \\
Operating Frequency ($f_c$) & 60 GHz \\
Access Point Tx Power ($P_t$) & 5 dBm \\
Deployment Area & $10 \times 10$ m$^2$ \\
Reflector arrays \& Users ($K$) & 2 arrays; $K \in \{2, 4\}$ \\
\\ 
\multicolumn{2}{c}{\textit{Policy \& Value Network Architecture}} \\
Manager Network (High-Level) & Attention + 128-unit FC (ReLU) \\
Agent Networks (Low-Level) & Two-layer FC (256 units, ReLU) \\
Optimizer & Adam \\
Learning Rate ($\eta$) & $2.0 \times 10^{-4}$ \\
\\ 
\multicolumn{2}{c}{\textit{MAPPO \& Training Configurations}} \\
Discount Factor ($\gamma$) \& GAE ($\lambda$) & 0.985, 0.9 \\
PPO Clipping Ratio ($\epsilon$) & 0.2 \\
Value Loss \& Entropy Coefficients & 1.0, $1.0 \times 10^{-4}$ \\
Optimization Epochs per Batch & 40 (Batch size = 200) \\
Total Training Episodes & 3,200 \\
Deployment Evaluation Horizon & 300 timesteps \\
Initial Focal Point Distribution & $\mathcal{N}([0, 0, 1.5]^\top \text{m}, 2.5\mathbf{I})$ \\
\hline
\end{tabular}
\end{table}

The MAPPO algorithm operates under a CTDE paradigm, where global system state information is accessible during training and at the centralized high-level controller, while individual low-level agents execute policies based solely on local observations during deployment. A complete summary of the system configuration, neural network architectures, and training hyperparameters is provided in Table~\ref{tab:system_params}.

\subsection{Convergence and Deployment Performance}

To evaluate the learning efficiency and practical deployment viability of the proposed framework, we analyze both training convergence and post-training signal stability in a 4-user scenario. The performance of the full hierarchical framework with geometric prior (\textit{HMARL-C}) is compared against a hierarchical variant without the compatibility matrix (\textit{HMARL-NC}), a conventional centralized PPO agent (\textit{Centralized}), and a random reflector--user allocation baseline (\textit{Random}).

\begin{figure}[!tbp]
    % \centering
    \subfloat[Training Convergence]{
        \includegraphics[width=0.95\linewidth]{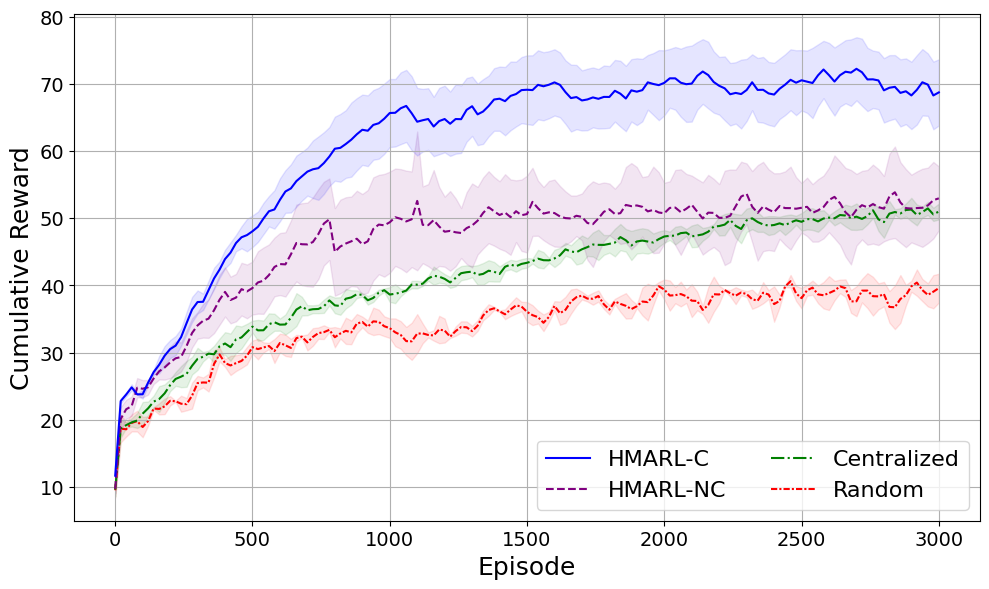}
        \label{fig:training_4ue}
    }
    \\
    \subfloat[Deployment RSSI]{
        \includegraphics[width=0.95\linewidth]{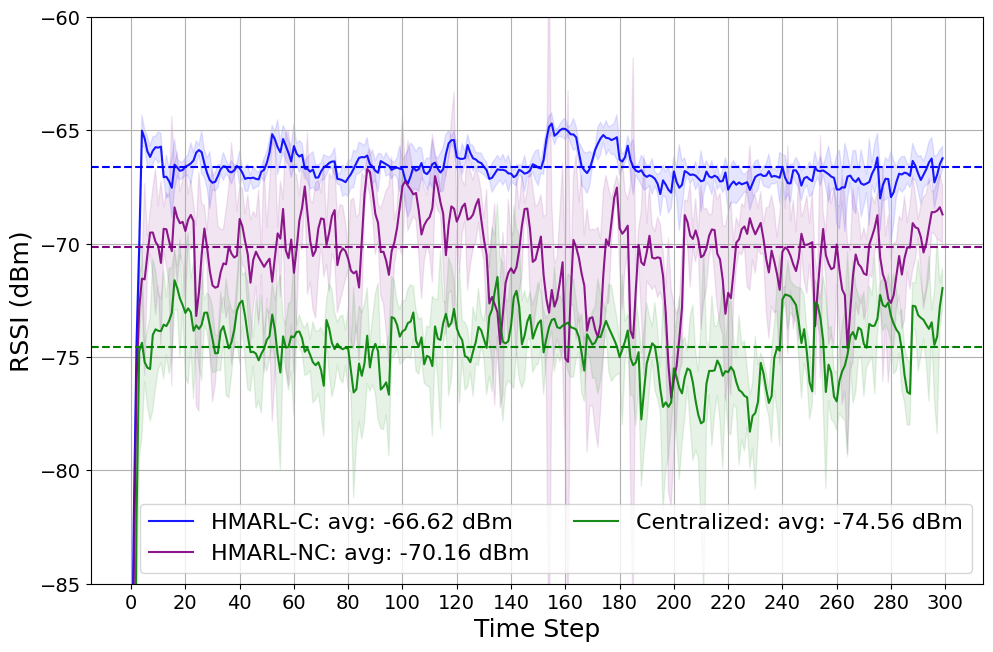}
        \label{fig:eval_4ue}
    }
    \caption{Performance evaluation for the 4-user configuration. (a) Cumulative reward convergence over 3,000 training episodes. (b) Deployment RSSI evaluated over 300 timesteps under continuous user mobility. Solid lines denote the mean, and shaded regions indicate the empirical standard deviation.}
    \label{fig:performance_4ue}
\end{figure}

\subsubsection{Training Convergence and the Inductive Bias}

Fig.~\ref{fig:training_4ue} illustrates the episode-averaged cumulative reward convergence over 3{,}000 training episodes. The \textit{HMARL-C} method exhibits rapid initial learning, separating from the baselines within the first 500 episodes and converging to an average reward of approximately 70. In contrast, \textit{HMARL-NC} plateaus around 51, the \textit{Centralized} agent stabilizes near 50, and the \textit{Random} baseline remains around 39.

The performance gap between \textit{HMARL-C} and \textit{HMARL-NC} isolates the contribution of the domain-specific compatibility matrix. In the massive combinatorial action space of a multi-user, multi-reflector environment, discovering near-optimal assignments through pure exploration is computationally expensive. The compatibility matrix provides an essential geometric inductive bias, guiding the high-level controller toward spatially favorable configurations and helping the agents avoid convergence to suboptimal local minima.

% Fig.~\ref{fig:training_4ue} illustrates the episode-averaged reward convergence over 3{,}000 training episodes. 
% The full \textit{Allocator} method exhibits rapid initial learning, breaking away from the baseline algorithms within the first 500 episodes and converging to a higher average reward of approximately 70. 
% In contrast, both the \textit{No\_compat} and \textit{No\_allocator} baselines plateau significantly lower, stabilizing near a reward of 50. Moreover, the random allocation of reflector-user (\textit{Random} baseline) does not achieve good performance and only reaches a cumulative reward of around 39.

% The performance gap between the \textit{Allocator} and the \textit{No\_compat} variant isolates the critical contribution of the domain-specific compatibility matrix. 
% In the massive combinatorial action space of a multi-user, multi-reflector environment, discovering near-optimal assignments through pure exploration is computationally expensive. 
% The matrix serves as an essential geometric inductive bias, guiding the high-level controller toward spatially favorable configurations and preventing the agents from converging into suboptimal local minima.

\subsubsection{Deployment RSSI and Hierarchical Improvement}

% To validate real-world operational stability, the trained policies are evaluated over 300 timesteps while introducing continuous user mobility with a velocity of \(1\,\text{m/s}\). 
% As shown in Fig.~\ref{fig:eval_4ue}, the hierarchical architecture achieves higher RSSI improvements over other methods.

% The full \textit{Allocator} maintains a mean RSSI of \(-66.47\,\text{dBm}\). 
% Conversely, the \textit{No\_allocator} baseline struggles with the expanded observation dimensionality and credit assignment complexity of simultaneous multi-reflector control, achieving only \(-74.26\,\text{dBm}\). 
% This demonstrates a \(7.79\,\text{dB}\) performance gain strictly attributable to the hierarchical decomposition. 
% Crucially, the hierarchical variant lacking the geometric prior (\textit{No\_compat}) achieves an intermediate performance of \(-70.59\,\text{dBm}\). While the hierarchical structure alone provides a \(\sim3.5\,\text{dB}\) advantage over the centralized baseline, it still underperforms the full \textit{Allocator} by over \(4.3\,\text{dB}\). This performance gap confirms the essential role of the compatibility matrix; without this geometric inductive bias, the high-level controller converges to suboptimal assignment policies that are less robust to continuous spatial changes.

To validate operational stability, the trained policies are evaluated over 300 timesteps while introducing continuous user mobility with a velocity of \(1\,\text{m/s}\). As shown in Fig.~\ref{fig:eval_4ue}, the hierarchical architecture achieves higher RSSI than the non-hierarchical baseline methods.

The \textit{HMARL-C} variant maintains a mean RSSI of \(-66.62\,\text{dBm}\), while \textit{HMARL-NC} achieves \(-70.16\,\text{dBm}\) and the \textit{Centralized} agent reaches only \(-74.56\,\text{dBm}\). This corresponds to a performance gain of roughly \(7.9\,\text{dB}\) for \textit{HMARL-C} over the centralized baseline, with \textit{HMARL-NC} providing an intermediate improvement of about \(4.4\,\text{dB}\). The hierarchy alone thus yields several dB gain, and the additional geometric prior further boosts RSSI and robustness under continuous spatial changes.

\begin{figure}[!t]
    \centering
    \includegraphics[width=0.95\columnwidth]{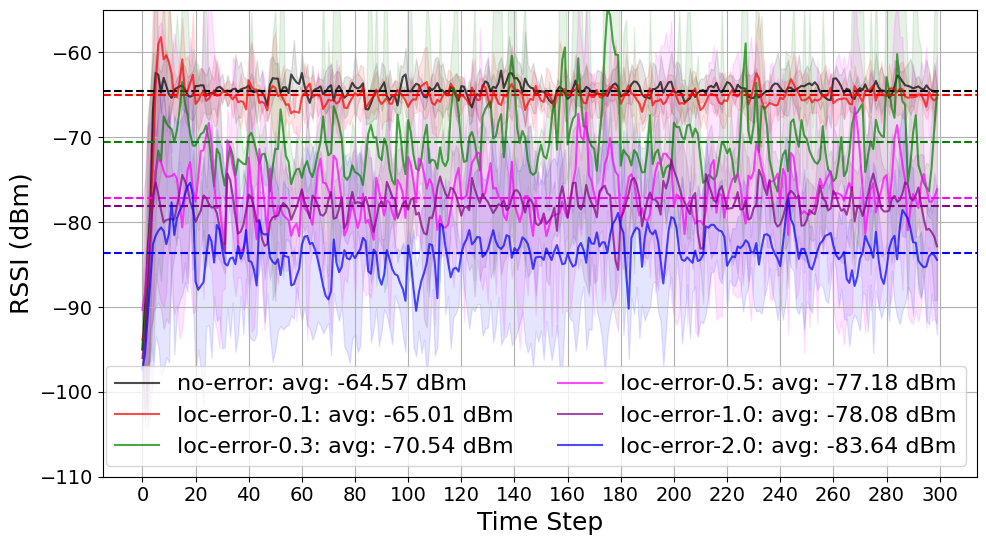}
    \caption{RSSI performance under varying degrees of user localization error for the 4-user configuration. The framework demonstrates robust, graceful degradation for sub-meter positioning noise, maintaining viable signal strength up to a 0.3 m error threshold. Solid lines represent the mean RSSI over 300 evaluation timesteps, with shaded regions denoting the empirical standard deviation.}
    \label{fig:localization_noise}
\end{figure}

\subsection{Sensitivity to Localization Errors}

Practical deployment of the proposed per-element CSI-free hierarchical framework depends on the availability of user localization information. 
Real-world localization systems inevitably encounter positioning errors due to hardware limitations and environmental multipath fading, which can degrade allocation quality and beam-focusing accuracy. 
To evaluate the framework's practical resilience, we simulate dynamic user tracking under varying localization error levels, modeled as a zero-mean Gaussian distribution with variance \(\sigma_{\text{error}} \in \{0.0, 0.1, 0.3, 0.5, 1.0, 2.0\}\) meters.
The agents are trained using error-matched statistics, reflecting practical deployments where tracking system tolerances are known a priori, thereby enabling error-aware policy learning.

As illustrated in Fig.~\ref{fig:localization_noise}, the system performance exhibits a systematic and graceful degradation corresponding with the tracking error magnitude. 
Under ideal conditions (no error), the 4-user system maintains an average RSSI of \(-64.57\,\text{dBm}\). 
When subjected to \(0.1\,\text{m}\) errors representative of emerging Ultra-Wideband (UWB) tracking systems, the system suffers a negligible penalty of roughly \(0.5\,\text{dB}\) (\(-65.01\,\text{dBm}\)). 
Operating within the \(0.3\,\text{m}\) error regime, which is typical of modern commodity WiFi or Bluetooth Low Energy (BLE) positioning infrastructure, the framework successfully secures a viable \(-70.54\,\text{dBm}\).

A critical operational boundary emerges at the \(0.5\,\text{m}\) threshold, where performance drops to \(-77.18\,\text{dBm}\) and the empirical standard deviation widens significantly. 
Errors exceeding \(1.0\,\text{m}\) lead to severe QoS degradation (from \(-78.08\,\text{dBm}\) to \(-83.64\,\text{dBm}\)) as the high-level controller misallocates reflectors and the continuous focal points miss their intended spatial targets. 
Nevertheless, by maintaining good sub-meter resilience, this evaluation confirms that the HMARL framework can be practically deployed using existing decimeter-level indoor tracking technologies. With the localization error analysis, we view the current design as most suitable for indoor or dense urban deployments where high-quality localization infrastructure is present.

\subsection{Limitations and Future Work}

While the proposed HMARL framework demonstrates high potential, the current evaluation is confined to deterministic ray-tracing simulations in specific indoor environments. First, because system performance degrades notably with localization errors exceeding 0.3 m, the framework's viability in outdoor scenarios with sparser tracking infrastructure remains an open challenge. Future work will investigate policy robustness under mismatched and non-Gaussian localization error distributions. Second, empirical validation using physical mechanical reflector prototypes and over-the-air measurements is necessary to capture real-world actuation latency and hardware imperfections. Moreover, subsequent studies will expand the evaluation scale to support massive multi-user, multi-reflector deployments.

\section{Conclusion}
\label{sec:conclusion}

This paper presents a HMARL framework to eliminate the prohibitive CSI estimation overhead in multi-reflector mmWave systems. By decomposing optimization into a high-level user allocator and low-level focal point controllers, the architecture leverages spatial localization and geometric priors instead of explicit per-tile CSI. Evaluations demonstrate this approach outperforms centralized baselines by up to \(7.9\,\text{dB}\) and maintains robust operation under typical sub-meter localization errors. Ultimately, hierarchical learning applied to mechanically reconfigurable reflectors provides a cost-effective, wideband alternative to electronic metasurfaces for next-generation wireless environments.

% This paper presents a HMARL framework to address the robustness and the CSI estimation overhead challenges inherent in multi-reflector mmWave systems. 
% By decomposing the optimization into high-level user allocation controller and low-level focal point controllers, the proposed architecture eliminates the dependency on explicit per-tile CSI, relying instead on spatial localization and geometric priors. 
% Experimental evaluations demonstrate that this approach outperforms centralized baselines by up to \(7.79\,\text{dB}\). 
% Crucially, the system exhibits practical deployment robustness, operating reliably typical sub-meter localization errors (\(\le 0.3\,\text{m}\)). 
% These findings establish mechanically reconfigurable reflector arrays, driven by hierarchical learning, as a highly viable, cost-effective, and wideband alternative to electronic metasurfaces for next-generation indoor wireless environments.

\bibliographystyle{IEEEtran}
\bibliography{IEEEabrv,./main_ref}

\end{document}